\documentclass[letterpaper, 10 pt, conference]{ieeeconf}  

\IEEEoverridecommandlockouts                              

\usepackage{graphics} 
\usepackage{epsfig} 
\usepackage{times} 
\usepackage{amsmath} 
\usepackage{amssymb}  
\usepackage{cite}

\usepackage{times}
\usepackage{amsfonts}
\usepackage{marvosym}
\usepackage{multicol}
\usepackage[bookmarks=true]{hyperref}
\usepackage{graphicx}
\usepackage{amsmath}
\usepackage{subcaption}
\usepackage{algorithm}
\usepackage{algorithmic}
\usepackage{color}
\usepackage{xcolor}
\usepackage{tcolorbox}
\usepackage{booktabs}
\usepackage{multirow}

\tcbuselibrary{breakable}
\tcbuselibrary{skins}
\let\labelindent\relax
\usepackage[inline]{enumitem}
\usepackage{listings}
\title{\LARGE \bf
Accelerating Robotic Reinforcement Learning with Agent Guidance
}

\author{Haojun Chen$^*$, Zili Zou$^*$, Chengdong Ma, Yaoxiang Pu, Haotong Zhang, Yuanpei Chen, Yaodong Yang
\thanks{Haojun Chen, Chengdong Ma, Haotong Zhang, Yuanpei Chen, and Yaodong Yang are with the Institute for Artificial Intelligence, Peking University. Additionally, Haojun Chen, Zili Zou, Yaoxiang Pu, Haotong Zhang, Yuanpei Chen, and Yaodong Yang are also with the PKU-PsiBot Joint Lab.
}
\thanks{*indicates equal contribution.}
\thanks{Corresponding to yaodong.yang@pku.edu.cn.}
}

\begin{document}

\maketitle
\thispagestyle{empty}
\pagestyle{empty}

\begin{abstract}

Reinforcement Learning (RL) offers a powerful paradigm for autonomous robots to master generalist manipulation skills through trial-and-error. However, its real-world application is stifled by low sample efficiency. Recent Human-in-the-Loop (HIL) methods accelerate training by using human corrections, yet this approach faces a scalability barrier. Reliance on human supervisors imposes a 1:1 supervision ratio that limits scalability, suffers from operator fatigue over extended sessions, and introduces high variance due to inconsistent human proficiency. We present Agent-guided Policy Search (AGPS), a framework that automates the training pipeline by replacing human supervisors with a multimodal agent. Our key insight is that the agent can be viewed as a semantic world model, injecting intrinsic value priors to structure physical exploration. By using tools, the agent provides precise guidance via corrective waypoints and spatial constraints for exploration pruning. We validate our approach on three tasks, ranging from precision insertion to deformable object manipulation. Results demonstrate that AGPS outperforms HIL methods in sample efficiency. This automates the supervision pipeline, unlocking the path to labor-free and scalable robot learning. \\
Project website: \url{https://agps-rl.github.io/agps/}.
\end{abstract}

\section{Introduction}
Deep Reinforcement Learning (RL) \cite{sutton1998introduction,rajeswaran2017learning} is a powerful paradigm for enabling autonomous robots to acquire general manipulation skills. RL allows agents to discover optimal policies without the constraints of hand-crafted modeling by leveraging trial-and-error interactions. However, the real-world application of RL is frequently hindered by low sample efficiency \cite{nakamoto2023cal, dulac2019challenges}.

\begin{figure*}[t]
    \centering
    \includegraphics[width=\linewidth]{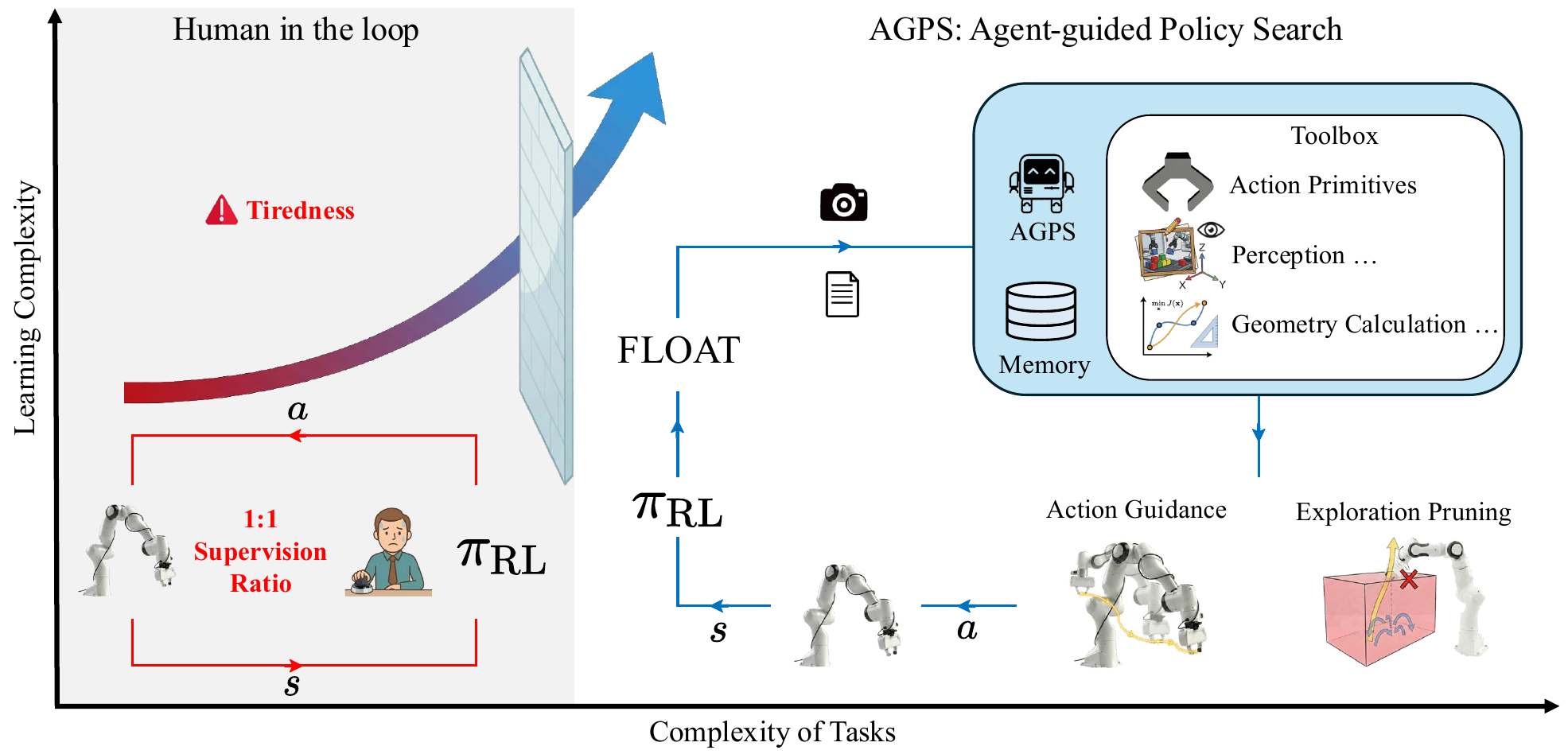}
    \caption{\textbf{Overview.} Left: HIL methods encounter a scalability barrier as task complexity rises, restricted by the 1:1 supervision ratio and operator fatigue. Right: AGPS transcends this barrier by automating supervision. The system employs FLOAT as an asynchronous trigger to monitor policy performance. When a deviation is detected, the agent recalls memory and leverages a toolbox (Action Primitives, Perception, Geometry) for spatial reasoning with an RGBD image and task description. These interventions manifest as Action Guidance for trajectory correction and Exploration Pruning for spatial constraining.}
    \label{fig:apgs}
    \vspace{-10pt} %
\end{figure*}

To speed up learning, Human-in-the-Loop (HIL) methods \cite{chen2025conrft, luo2025precise} are commonly used to guide robots. This approach works well for single tasks. However, it faces a ``scalability barrier,'' as shown in Figure \ref{fig:apgs}. This barrier arises from a mismatch. Robot tasks become more complex, creating more situations to handle. But humans can only provide limited guidance. Each robot needs one person to supervise it. This makes it hard to expand to many robots. In addition, humans become tired during long training processes. Their guidance becomes less accurate and slower. Therefore, the need for supervision grows beyond what humans can provide. This prevents HIL methods from scaling to handle multiple tasks.

To overcome the limitations depicted in Figure \ref{fig:apgs}, we need guidance that does not depend on human labor. We propose using multimodal agents. These agents can be viewed as semantic world models, injecting intrinsic value priors derived from internet-scale pretraining into the robotic system beyond offering scalable supervision. The agents can use tools to apply this knowledge. By calling algorithms for visual grounding and spatial calculation, the agents can identify task-relevant regions, thus pruning the search space. In this work, we present \textbf{Agent-guided Policy Search (AGPS)}, a framework that uses these capabilities to automate robot training. To bridge the frequency gap between high-speed RL interactions and low-frequency agent reasoning, we incorporate FLOAT \cite{yu2025armada}, an online failure detector. FLOAT monitors the policy’s behavior in real-time and triggers the agent only when distribution drift occurs. Upon activation, the agent uses tools to provide supervision in two ways: (1) Action Guidance, which generates correct waypoints to recover from failure states, and (2) Exploration Pruning, which defines 3D spatial constraints to mask out task-irrelevant states.

We evaluate AGPS on three challenging real-world manipulation tasks spanning distinct physical properties: USB insertion, which demands high-precision control for rigid body assembly, and Chinese knot hanging, which requires intricate interaction with deformable linear objects, and towel folding, which involves the complex manipulation of high-dimensional, deformable surfaces. Our results demonstrate that AGPS outperforms the HIL methods in sample efficiency with zero human intervention. In summary, this work makes two main contributions. \textbf{First,} we propose AGPS, a framework that automates RL supervision by integrating a multimodal agent with the FLOAT trigger mechanism to reduce the inference cost. \textbf{Second,} we demonstrate that AGPS significantly outperforms baselines without human in the loop through real-world experiments.

\section{Related Work}
\noindent\textbf{Real-world RL.}
Improving the sample efficiency of real-world robotic reinforcement learning (RL) has involved several key strategies: developing highly efficient off-policy algorithms \cite{ball2023efficient, luo2024serl}; leveraging demonstrations and prior data through hybrid methods \cite{rajeswaran2017learning, hu2023imitation, nakamoto2023cal}, or guided curricula \cite{uchendu2023jump}; employing residual learning architectures \cite{johannink2019residual, ankile2025residual, yuan2024policy, dong2025expo, xiao2025self} to refine base policies. SimLauncher \cite{wu2025simlauncher} bootstraps real-world RL using simulated rollouts and action proposals. To further accelerate exploration, HIL methods like HIL-SERL \cite{luo2025precise} and ConRFT \cite{chen2025conrft} leverage online human corrections to prune the search space and ensure safety, akin to the constrained trajectory optimization in GPS \cite{levine2013guided}. However, HIL approaches are bounded by a strict 1:1 supervision ratio. In contrast, our work replaces human with a multimodal agent, providing consistent guidance that may scale up without human intervention.

\noindent\textbf{Foundation models for policy learning.}
Foundational models are transforming robotic policy learning across multiple paradigms. Early works like SayCan \cite{ahn2022can} and Inner Monologue \cite{huang2022inner} used Large Language Models (LLMs) for semantic planning, later extended by Code as Policies \cite{10160591}, ProgPrompt \cite{singh2022progprompt}, and VoxPoser \cite{huang2023voxposer} for program synthesis. For direct perception-to-action mapping, \cite{brohan2022rt, zitkovich2023rt, driess2023palm, jiang2022vima} demonstrated large-scale visuomotor control. To automate reward design, Eureka \cite{ma2024eureka} and ReKep \cite{pmlr-v270-huang25g} leverage LLMs or Vision Language Models (VLMs) to generate reward functions or cost constraints. Foundation models also provide priors for RL, as in RLFP\cite{pmlr-v270-ye25a}, and scale data via automated demonstration generation for policies \cite{duan2025manipulate, ha2023scaling, goyal2024rvt}. Visual grounding is advanced in \cite{liu2024moka, villa2023pivot, pmlr-v270-yuan25c}, while spatial reasoning is studied in \cite{Hsu_2023_CVPR, Chen_2024_CVPR}. Additional contributions include visual priors \cite{ma2023vip, pmlr-v205-nair23a}, exploration \cite{wang2024voyager}, unified policies \cite{NEURIPS2023_1d5b9233}, and instruction conditioning \cite{pmlr-v164-shridhar22a}.

\section{Preliminaries}
This section introduces the reinforcement learning framework and Optimal Transport (OT) \cite{peyre2019computational} for failure detection.

\subsection{Reinforcement Learning}
We model the manipulation task as a finite-horizon Markov Decision Process (MDP), denoted by $\mathcal{M} = (\mathcal{S}, \mathcal{A}, \mathcal{P}, \mathcal{R}, \gamma, H)$. The policy $\pi_\theta(a_t|s_t)$ maps states $s_t$ (RGB images and proprioception) to actions $a_t$. The environment transitions according to $\mathcal{P}(s_{t+1}|s_t, a_t)$ and provides a reward $r_t = \mathcal{R}(s_t, a_t)$. The goal is to maximize the expected discounted return $J(\pi) = \mathbb{E}_{\pi} [\sum_{t=0}^{H} \gamma^t r_t]$. We use expert demonstrations $\mathcal{D}_E$ as references for policy deviation.

\subsection{Optimal Transport for Trajectory Matching}
We use OT as a metric to quantify how current policy rollouts deviate from the expert behaviors. OT provides a geometric distance between distributions that is resilient to temporal shifts. We map observations from expert trajectories $\mathcal{T}_e$ and agent rollouts $\mathcal{T}_b$ into a latent space using a pretrained encoder $\phi(\cdot)$, yielding embeddings $X_e$ and $X_b$. The OT distance $d_{\text{OT}}(X_e, X_b)$ is the minimum cost to transport mass between these distributions:$$d_{\text{OT}}(X_e, X_b) = \min_{\mu} \sum_{i,j} C_{ij} \mu_{ij}$$subject to marginal constraints $\sum_{j} \mu_{ij} = \frac{1}{L_e}$ and $\sum_{i} \mu_{ij} = \frac{1}{L_b}$. The cost matrix $C_{ij} = 1 - \cos(\phi(o_{e,i}), \phi(o_{b,j}))$ measures embedding dissimilarity. This defines the FLOAT Index, $\lambda(\mathcal{T}_b) = \min_{n} d_{\text{OT}}(\phi(\mathcal{T}_e^n), \phi(\mathcal{T}_b))$, measuring the deviation from expert distribution.

\section{Agent-guided Policy Search}
We propose Agent-guided Policy Search (AGPS), a framework illustrated in Figure \ref{fig:apgs}. It consists of two core components: the asynchronous failure detector (FLOAT) that monitors the policy $\pi_{\text{RL}}$, and a toolbox that translates semantic understanding into corrective waypoints and spatial constraints. The algorithm pseudocode is provided in Appendix \ref{sec:code}.

\subsection{Asynchronous Failure Detection (FLOAT)}
The multimodal agents have high inference latency, making them unsuitable for high-frequency robotic control. To mitigate the high inference latency associated with agents, we employ FLOAT as a real-time trigger. This module monitors policy execution and solicits guidance only when a significant deviation from the expert distribution is detected.

We utilize a pre-trained visual encoder $\phi(\cdot)$ (DINOv2 ViT-B/14 \cite{oquabdinov2}) to project observations into a latent feature space. At timestep $t$, the system calculates a deviation score $\lambda_t$ based on the OT distance between the current rollout $\mathcal{T}_{b}$ and the set of expert demonstrations $\mathcal{T}_{e}$:
\begin{equation}
\lambda_t = \min_{n \in \{1,\dots,N\}} d_{\text{OT}}(\phi(\mathcal{T}_e^n), \phi(\mathcal{T}_{b, 1:t}))
\label{eq:lambda_t}
\end{equation}

We set the threshold $\Lambda$ to the 95th percentile of $\{\lambda_e^n\}_{n=1}^N$ from expert demonstrations, where $\lambda_e^n = d_{OT}(\phi(\mathcal{T}_e^n), \phi(\mathcal{D}_E \setminus \{\mathcal{T}_e^n\}))$. If $\lambda_t \le \Lambda$, $\pi_\text{RL}$ continues execution to maintain high throughput. If $\lambda_t > \Lambda$, the system pauses and queries the agent for corrective guidance.

\subsection{The toolbox} 
The primary challenge in this pipeline is grounding semantic knowledge in the physical world. We address this by developing a toolbox of executable modules that allow the agent to reason about the workspace with geometric precision.

\subsubsection{Perception Module}
The agent utilizes a VLM to identify task-relevant keypoints (e.g., ``USB port'', ``socket'') from RGBD images. The VLM outputs 2D pixel coordinates $\mathbf{u}_{\text{key}}$, which are then deprojected into 3D world coordinates $P_{\text{world}}$ by Eq. \ref{eq:deproject},
\begin{equation}
    P_{\text{world}} = R^{-1} \left( D(\mathbf{u}_{\text{key}}) \cdot K^{-1} [\mathbf{u}_{\text{key}}, 1]^T - \mathbf{t} \right)
    \label{eq:deproject}
\end{equation}
where $K \in \mathbb{R}^{3 \times 3}$ is the camera intrinsic matrix, and $[R, \mathbf{t}] \in SE(3)$ represents the camera extrinsic parameters relative to the robot base. This transformation maps semantic features from the image plane into the robot's task space, providing a metric foundation for geometric guidance.

\subsubsection{Action Primitives Library}
We define a library of atomic Action Primitives to serve as waypoint generators. Based on the current TCP pose and gripper state, these primitives calculate specific target configurations. They include discrete actions (e.g., Grasp, Release) and continuous motions (e.g., MoveDelta, Lift). This modular design allows the agent to compose precise geometric interventions efficiently.

\subsubsection{Memory Module}
To reduce system latency, an episodic memory module caches spatial constraints, such as bounding boxes $\mathcal{C}_{\text{box}}$. Before invoking the VLM, the system queries this memory; if a historical rollout was successful, the agent reuses the stored bounding box to avoid redundant VLM inference.

\subsection{Automated Guidance Mechanisms} Leveraging the toolbox, AGPS provides two types of interventions:
\subsubsection{Action Guidance}
Upon a FLOAT trigger, the agent identifies the failure mode and synthesizes a corrective trajectory. By selecting appropriate action primitives and grounding them through the Geometry Module, the agent generates precise corrective waypoints. This mechanism provides a stable supervision signal that helps the policy not stray too far from the expert distribution. 
\subsubsection{Exploration Pruning}
To accelerate learning, the agent defines a 3D bounding box $\mathcal{C}_{\text{box}}$ that encapsulates the task-relevant volume. During RL training, actions leading outside this volume are masked. This constrains the search space to a valid region, preventing the robot from exploring irrelevant states.

\begin{figure*}[t]
    \centering
    \begin{minipage}[b]{0.23\textwidth}
        \centering
        \includegraphics[width=\textwidth]{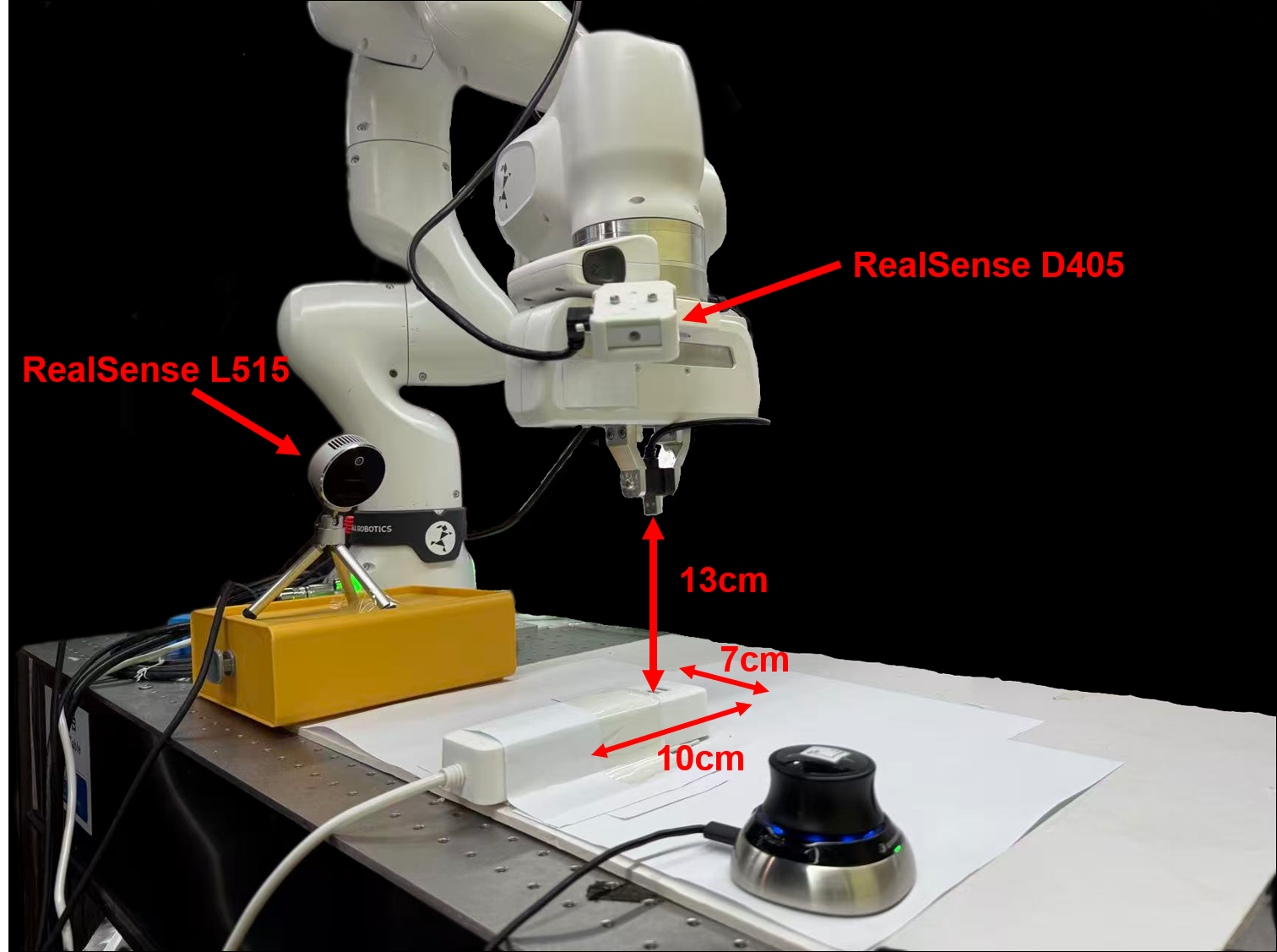}
        \centerline{\textbf{(a) USB Insertion}}
    \end{minipage}
    \hfill
    \begin{minipage}[b]{0.23\textwidth}
        \centering
        \includegraphics[width=\textwidth]{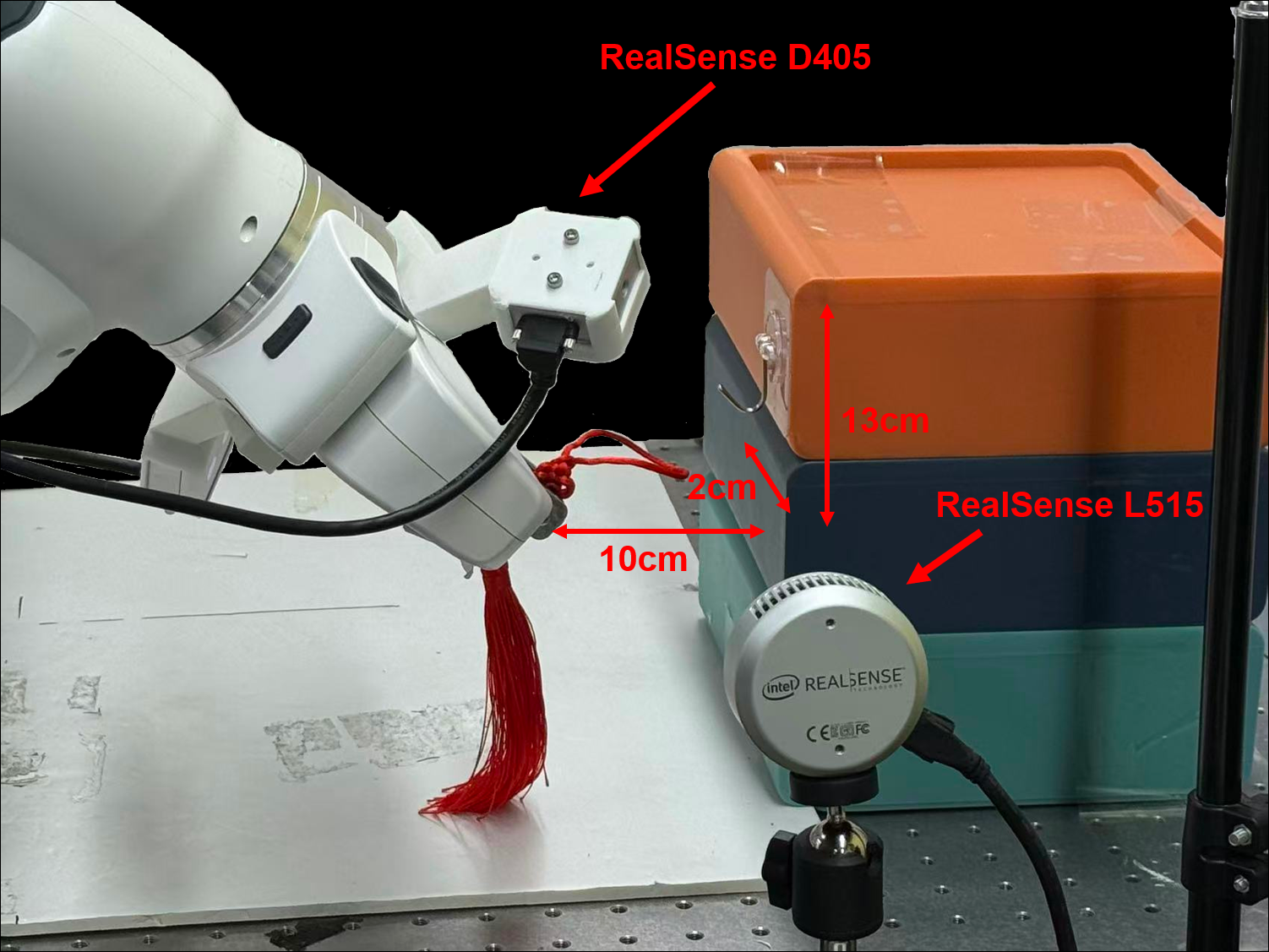}
        \centerline{\textbf{(b) Chinese Knot Hanging}}
    \end{minipage}
    \hfill
    \begin{minipage}[b]{0.23\textwidth}
        \centering
        \includegraphics[width=\textwidth]{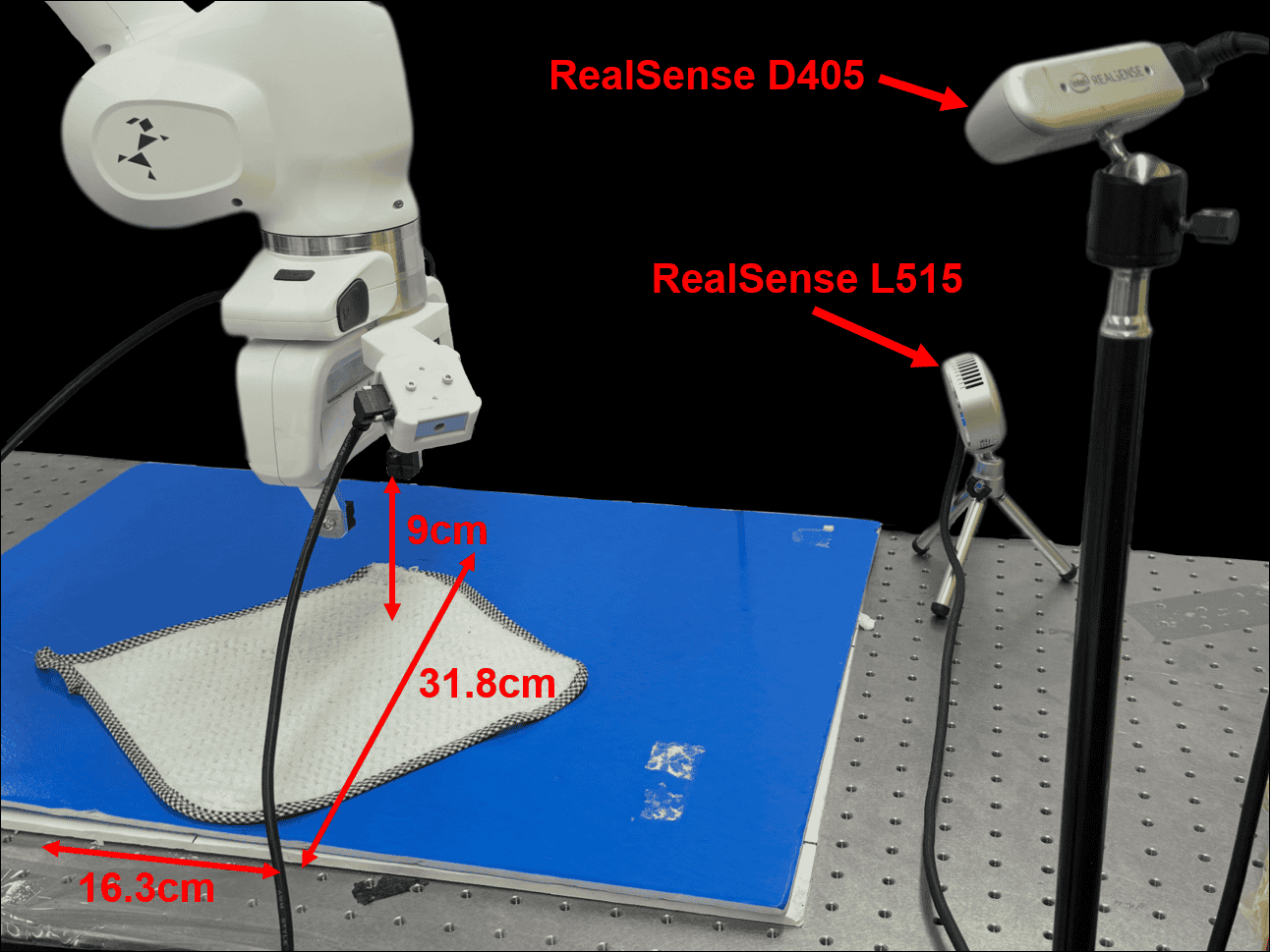}
        \centerline{\textbf{(c) Towel Folding}}
    \end{minipage}
    \hfill
    \begin{minipage}[b]{0.23\textwidth}
        \centering
        \includegraphics[width=\textwidth]{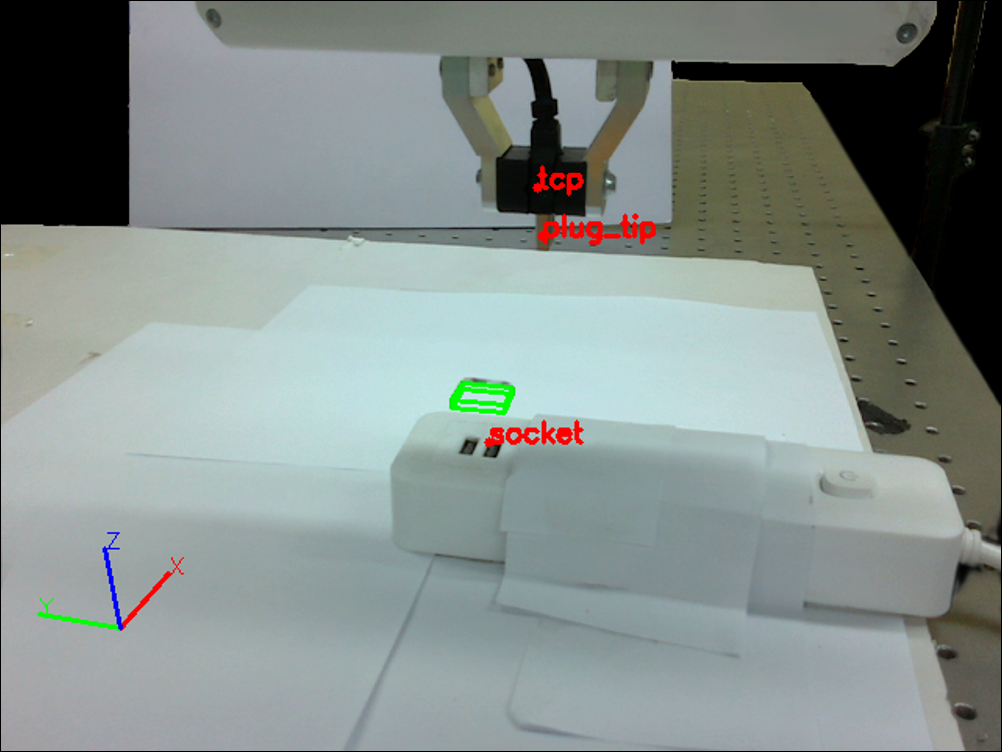}
        \centerline{\textbf{(d) Spatial Reasoning}}
    \end{minipage}
    
    \vspace{0.2cm}
    \caption{\textbf{Experimental Setup and AGPS Spatial Reasoning.} (a)-(c) show the hardware setup for three manipulation tasks, where red arrows and annotations indicate cameras and the reachable workspace of the Franka gripper. (d) illustrates the \textbf{AGPS spatial reasoning} capability, where red points denote keypoints and the green box represents the task-relevant volume for exploration pruning.}
    \label{fig:experiment_workspace_and_usb_bbox}
    \vspace{-10pt} %
\end{figure*}

\section{Experiments} 
Our experiments aim to address four questions: (1) Can AGPS significantly improve sample efficiency (Section \ref{sec:overal_performance})? (2) Does the frequency of agent interventions decay as the policy improves (Section \ref{sec:intervention_reduce})? (3) What is the quality of AGPS's intervention (Section \ref{sec:policy-training-dynamics})? (4) Does memory accelerate training (Section \ref{sec:memory})?

\subsection{Experimental Setup}
We evaluate AGPS on a diverse suite of real-world manipulation tasks spanning rigid and deformable object domains. Following prior work \cite{luo2024serl,luo2025precise,chen2025conrft}, we train a binary classifier as the reward function for each task. The experimental workspaces and initialization ranges are illustrated in Figure \ref{fig:experiment_workspace_and_usb_bbox}a-c. Details are provided in Appendix \ref{sec:train-details}.

\noindent\textbf{USB Insertion:} The robot aligns and inserts a USB connector into a charging slot, demanding sub-millimeter accuracy. The gripper starts with the connector grasped. The action space is 6-dimensional (delta end-effector pose), with the initial pose randomized within $\pm 6$cm (XY), $\pm 5$cm (Z), and $\pm 0.06$rad (RZ). The reward is a sparse binary signal with one on success, zeros for the other steps.

\noindent\textbf{Chinese Knot Hanging:} The robot hangs a deformable Chinese Knot onto a hook. The knot sags under gravity and deforms during motion, requiring the robot to approach from above. The gripper starts with the knot string grasped. The action space is 6-dimensional, with the initial position randomized within $\pm 6$cm (XY) and $\pm 0.06$rad (RZ). The reward is a sparse binary signal with one on success, zeros for the other steps.

\noindent\textbf{Towel Folding:} Unlike the previous tasks, the gripper begins open. The robot must first grasp one corner of the towel, then fold it toward the opposite corner on a smooth cardboard surface where any gripper contact displaces the towel. The initial position is randomized within $\pm 5$cm (XY) and $\pm 0.02$rad (RZ). The action space is 7-dimensional (6-DoF delta pose and a binary gripper command). Due to the larger action space and multi-stage nature, the reward is $r=10$ on success and $r=-0.1$ per step otherwise.

\subsection{Baselines and Metrics}
Since AGPS is a plug-and-play framework independent of RL algorithm, we evaluate it with different SOTA RL algorithms across tasks. For USB Insertion and Chinese Knot Hanging, we compare AGPS against SERL \cite{luo2024serl} (off-policy RL with offline demonstrations), HIL-SERL \cite{luo2025precise} (SERL with online human interventions). For Towel Folding, where multi-stage deformable manipulation is difficult to learn from scratch, we adopt the two-stage reinforcement fine-tuning paradigm. In the first stage, the policy is pre-trained on 40 expert trajectories for 48,000 training steps by CaL-ConRFT \cite{chen2025conrft}. In the second stage, the policy is fine-tuned via online RL. We replace the human intervention component in the original HIL-ConRFT \cite{chen2025conrft} with our AGPS framework. We compare AGPS against HIL-ConRFT \cite{chen2025conrft} under identical network architectures and hyperparameters following the official implementation.

The performance is measured by three metrics. Success Rate tracks completion probability over 10 randomized trials. Time to Convergence records the wall-clock time to reach 100\% success. Intervention Ratio measures the frequency of external guidance or agent activations per episode.

\subsection{Overall Performance}
\label{sec:overal_performance}
\begin{figure*}[h]
    \centering
    \includegraphics[width=\linewidth]{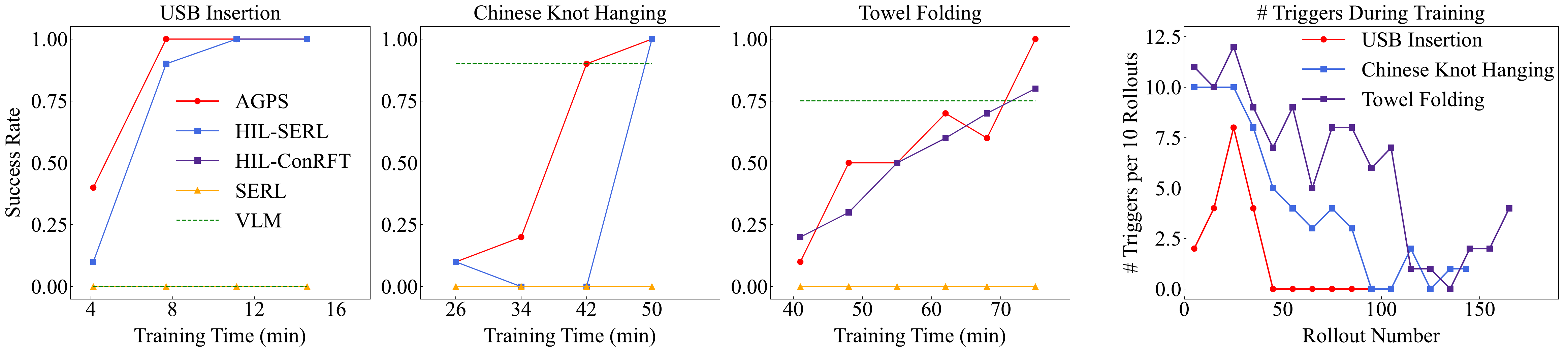}
    \caption{\textbf{Overall performance comparisons and Intervention Analysis.} (a-c) Success rates of AGPS compared to baselines across three tasks. AGPS achieves higher sample efficiency and final success rates. (d) The number of triggers per 10 rollouts decreases over time in all tasks.}
    \label{fig:overall_perf_and_float}
\end{figure*}

AGPS demonstrates superior sample efficiency across all tasks. In USB Insertion task(Figure \ref{fig:overall_perf_and_float}a), it achieves 40\% success at 4 minutes and converges to 100\% at 8 minutes, outperforming HIL-SERL. This gain stems from Exploration Pruning. As shown in Figure \ref{fig:experiment_workspace_and_usb_bbox}d, AGPS restricts the gripper to a task-relevant bounding box, preventing wasted exploration. SERL fails (0\%) due to the vast search space.

In the Chinese knot hanging task (Figure \ref{fig:overall_perf_and_float}b), 
AGPS converges significantly faster than HIL-SERL. HIL-SERL remains at 0\% success rate until 42 minutes, hampered by inconsistent human interventions common in deformable object manipulation. In contrast, AGPS reaches 90\% success at 42 minutes and 100\% at 50 minutes. This indicates that consistent agent-generated waypoints provide more effective guidance than high-variance human teleoperation. SERL remains at 0\%.

In the towel folding task (Figure \ref{fig:overall_perf_and_float}c), both methods initially suffer from policy collapse, where the gripper moves directly to the target corner without grasping the towel. Continuous external guidance is required to correct this behavior. As training progresses, both AGPS and HIL-ConRFT gradually improve. However, AGPS ultimately outperforms HIL-ConRFT, achieving higher performance without human intervention. This stems from the agent's ability to provide sustained, low-variance interventions, avoiding the inconsistency and fatigue inherent to prolonged human supervision.

\subsection{Intervention Frequency}
\label{sec:intervention_reduce}
We tracked the number of VLM triggers during training (Figure \ref{fig:overall_perf_and_float}d). In the USB Insertion (Red) task, the intervention rate peaks early. It drops to 0 by Rollout 45. The policy $\pi_{\text{RL}}$ quickly masters the insertion. In the Chinese Knot (Blue) task, guidance starts at the maximum (10 triggers). It decreases slowly due to complex rope dynamics. Interventions reach near zero around Rollout 100. In the towel folding task (purple), the policy is prone to collapse due to the high-dimensional state space, causing the intervention rate to oscillate significantly. Despite these fluctuations, the overall trend remains downward, indicating that $\pi_{\text{RL}}$ progressively internalizes the agent's guidance and learns to handle the complex dynamics autonomously.

\begin{figure}[h]
    \centering
    \begin{subfigure}[t]{0.23\textwidth}
        \centering
        \includegraphics[height=3.0cm]{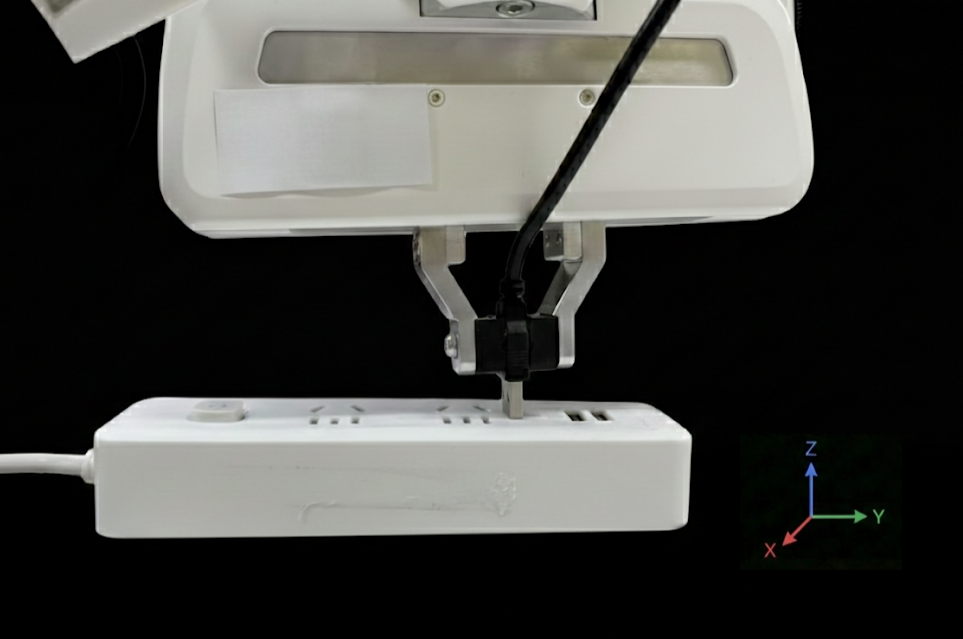}
        \caption{\textbf{Failure case of HIL-SERL.}}
        \label{fig:usb-failure-case}
    \end{subfigure}
    \hfill
    \begin{subfigure}[t]{0.24\textwidth}
        \centering
        \includegraphics[width=\linewidth]{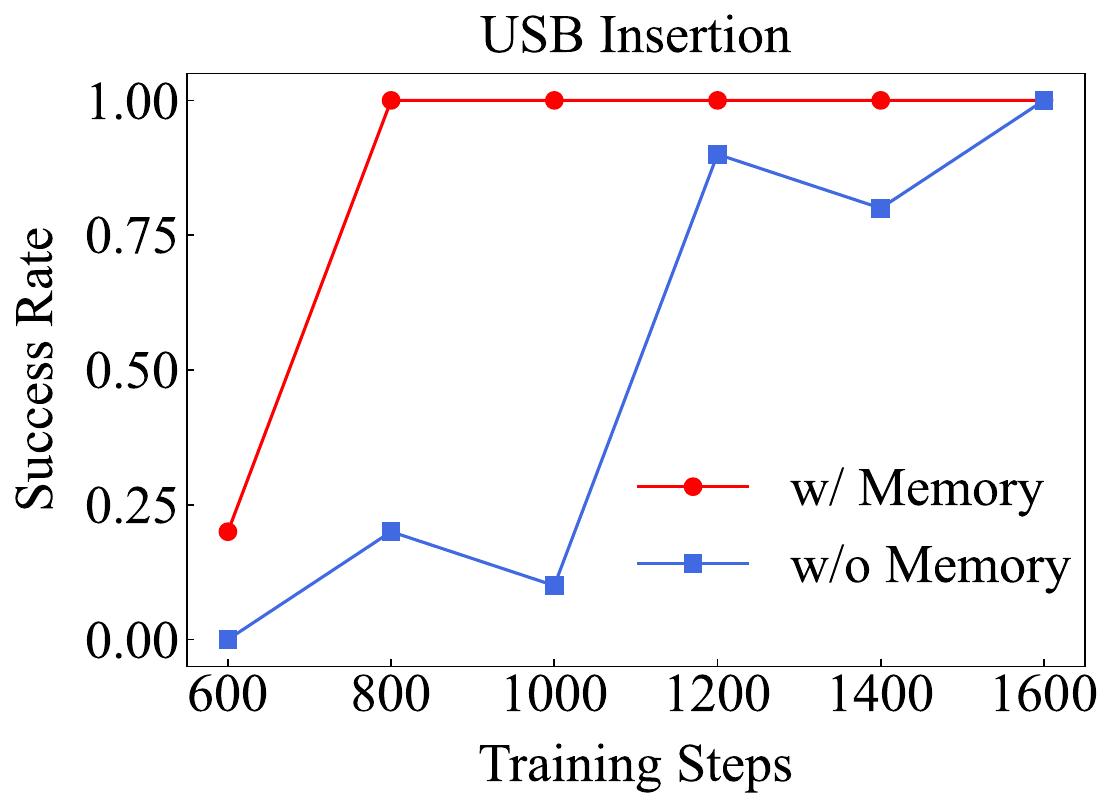}
        \caption{\textbf{Ablation on memory.}}
        \label{fig:memory}
    \end{subfigure}
    \caption{\textbf{Failure case and Ablation.} (a) illustrates a failure case. (b) shows that the memory module (red) accelerates convergence by $2\times$ compared to the baseline (blue).}
    \label{fig:usb-failure-case-and-abla-memory}
    \vspace{-10pt} %
\end{figure}

\subsection{Generalization and State-Value Analysis}
\label{sec:policy-training-dynamics}
\begin{figure}[t]
    \centering
    \includegraphics[width=\linewidth]{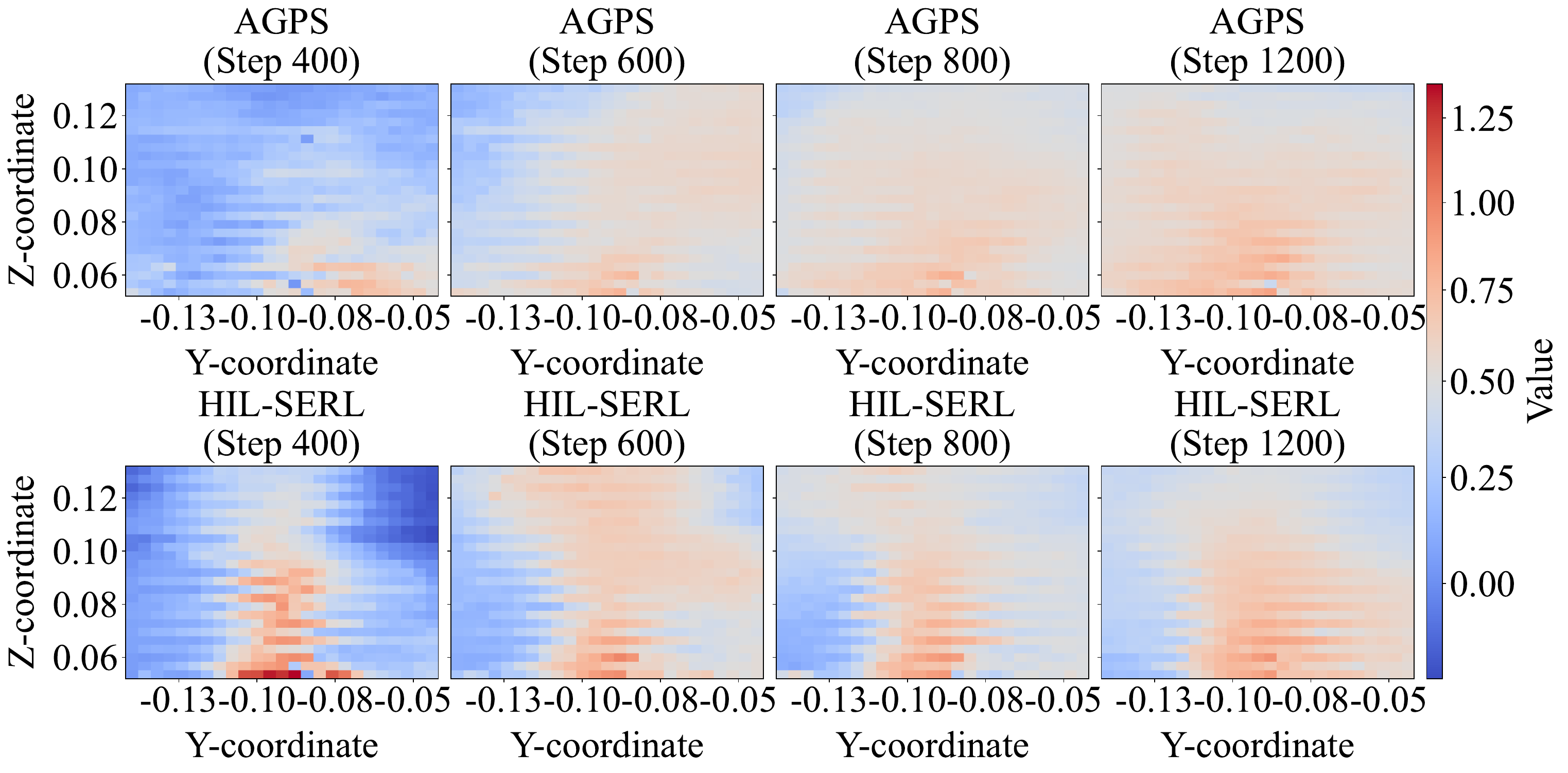}
    \caption{\textbf{Visualization of policy training dynamics.} Value distributions across Y-Z spatial coordinates at different training steps. (Bottom) HIL-SERL learns a narrow high-value corridor, indicating overfitting to human demonstrations, leaving surrounding states with near-zero value (blue regions). (Top) AGPS develops a broad high-value funnel, showing learned recovery behaviors for states deviating from the expert distribution.}
    \label{fig:q_value}
    \vspace{-10pt} %
\end{figure}

We analyze the learned state-value landscapes to investigate how exploration strategies affect policy generalization (Figure \ref{fig:q_value}). HIL-SERL (bottom row) forms a narrow high-value corridor because human operators demonstrate only the most direct paths. This bias causes the policy to overfit to a thin trajectory, leaving off-center states with near-zero values (blue regions). Conversely, AGPS (top row) develops a broader high-value landscape. Since the FLOAT trigger only intervenes during critical failures, the policy autonomously resolves minor misalignments, forcing it to learn recovery behaviors. This value distribution directly impacts physical performance. In a test case (Figure \ref{fig:usb-failure-case}), HIL-SERL consistently fails. Lacking gradient information in this low-value region ($Y \approx -0.13$), the robot freezes. AGPS succeeds because its broad landscape covers that state, providing the necessary gradients to execute corrective shifts toward the slot. This confirms that agent-guided exploration enables the policy to recover from diverse initial states.

\subsection{Memory accelerates training}
\label{sec:memory}
We conducted an ablation study on the USB task to evaluate the memory module (Figure \ref{fig:memory}). AGPS with Memory (Red) converges by Step 800, achieving a 2$\times$ speedup over the version without memory (Blue), which requires 1600 steps. This acceleration results from reusing validated spatial constraints. By retrieving successful bounding boxes from history, the system bypasses redundant VLM computations. Consequently, the policy converges significantly faster by avoiding repeated calls to the agent.

\subsection{Failure modes of AGPS}
\label{sec:vlm_only}
We evaluate the standalone capabilities of the Action Guidance module. The VLM first identifies task-relevant keypoints and then queries the agent to generate action primitives based on these spatial references. We conduct 10 trials per task, with success rates plotted as green dashed lines in Figures \ref{fig:overall_perf_and_float}a-c. The agent achieved 0\% success for USB insertion, 90\% for Chinese Knot Hanging, and 70\% for towel folding. We observe two primary failure modes. First, \textit{imprecise localization} impedes high-precision tasks. In the USB Insertion task, the VLM fails completely (0\%) because perception noise prevents the sub-millimeter accuracy required for rigid assembly. Similarly, in the Chinese Knot Hanging, the agent occasionally mislocalizes the thin string, targeting the table surface instead. Second, \textit{hallucination} occurs in complex scenes; for Towel Folding, the VLM struggles to robustly identify keypoints when the fabric is crumpled or self-occluded.

Despite these imperfections, AGPS generates correct guidance in most cases. Suboptimal trajectories serve as valuable negative examples. The RL policy learns to avoid these failed attempts through low rewards, while successful guidance accelerates training. Thus, while VLMs are insufficient as standalone controllers for precision tasks, they serve as effective teachers for RL.

\section{Discussion: AGPS as a Semantic World Model}
\label{sec:discussion}
\begin{figure}
    \centering
    \includegraphics[width=0.8\linewidth]{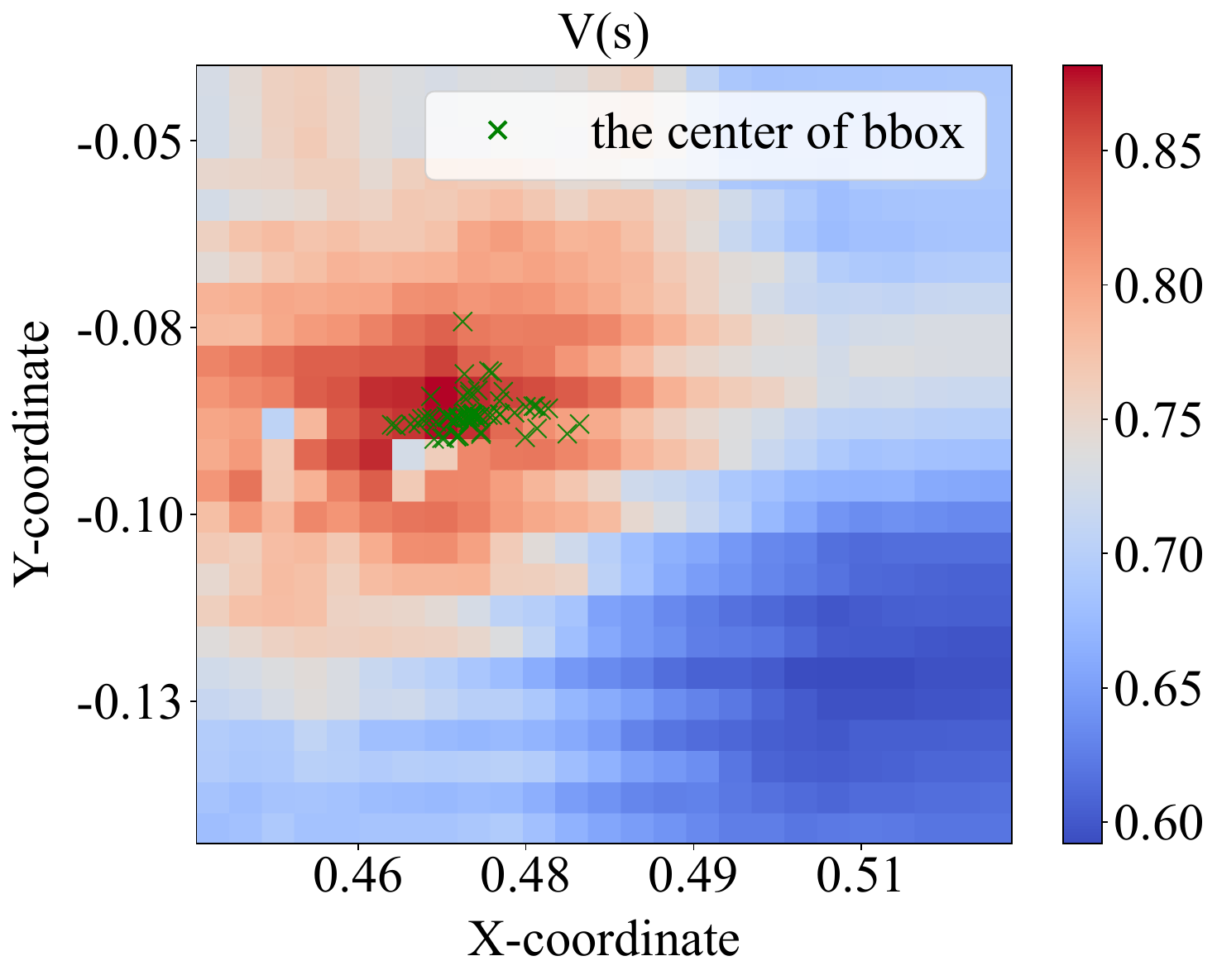}
    \caption{\textbf{Visualization of spatial alignment.} The heatmap displays the learned state-value distribution of the converged HIL-SERL policy, where red indicates high value. Green crosses mark the centers of bounding boxes generated by AGPS. Their tight alignment with the high-value region confirms that the agent's semantic priors accurately predict the high value region without prior training.}
    \label{fig:q-value-alignment}
    \vspace{-10pt}
\end{figure}

Real-world reinforcement learning is limited by expensive, serial interactions. Our results suggest that multimodal agents can be viewed as semantic world models to address this bottleneck. This is evidenced by the spatial alignment shown in Figure \ref{fig:q-value-alignment}. The heatmap displays the learned state-value landscape of HIL-SERL. Green crosses, marking zero-shot bounding box centers generated by our agent, align closely with the high value regions (red).

While the baseline requires human intervention to find the high value region, the agent identifies it immediately through semantic reasoning. This suggests that foundation models possess an intrinsic value prior, predicting the high-value region without interaction. By using this prior to prune the exploration space, AGPS allows the RL algorithm to focus on task-relevant regions. This approach replaces human supervision with autonomous semantic constraints, enabling more scalable real-world learning.

\section{limitations}

Our framework faces limitations primarily regarding the VLMs. Visual grounding errors can lead to intervention failures. Furthermore, high inference latency limits the frequency of guidance, restricting applicability in highly dynamic scenarios. However, we anticipate that advancements in VLM spatial reasoning will enable AGPS to supervise complex tasks, facilitating scalable learning.
A second limitation is the reliance on manual resets, particularly in towel folding. While outside the scope of this study on sample efficiency, AGPS could be extended to learn reset policies similar to \cite{hu2023rebootreusedatabootstrapping}, such as towel unfolding. We posit that accumulating such a library of action skills is essential for achieving fully automated real-world reinforcement learning.

\section{Conclusion} 
\label{sec:conclusion}
In this work, we introduced Agent-guided Policy Search (AGPS). This framework automates real-world reinforcement learning by replacing human supervision with a multimodal agent. By integrating asynchronous failure detection with an executable toolbox, the system grounds high-level semantic reasoning into precise physical constraints. Real-world experiments on rigid and deformable object manipulation demonstrate that AGPS outperforms the sample efficiency of HIL methods. Crucially, it achieves this with zero human intervention. Our analysis further reveals that the agent can be viewed as a pre-trained semantic world model. It aligns exploration with high-value task manifolds through zero-shot spatial pruning. These findings suggest a fundamental shift for real-world robotic learning. The path to scalability lies in leveraging agents to structure physical exploration, effectively replacing unscalable human labor with robust, autonomous semantic priors.

\section*{APPENDIX}
\renewcommand{\thesubsection}{\Alph{subsection}}
\setcounter{subsection}{0}
\subsection{Pseudocode of AGPS}
\label{sec:code}

In Algorithm \ref{alg:agps}, we present the detailed pseudocode of Agent-guided Policy Search (AGPS). 

\begin{algorithm}[h]
    \caption{Agent-guided Policy Search (AGPS)}
    \label{alg:agps}
    \begin{algorithmic}[1]
        \REQUIRE Pre-trained VLM Agent $\mathcal{A}_{\text{vlm}}$, Policy $\pi_{\theta}$, Critic $Q_{\phi}$. Expert dataset $\mathcal{D}_{E}$. Batch size $B$.
        \STATE Online buffer $\mathcal{R} \gets \emptyset$, Episodic Memory $\mathcal{M} \gets \emptyset$.
        \STATE Compute failure threshold $\Lambda$ based on $\mathcal{D}_E$.
        \STATE Randomly initialize $\pi_{\theta}$ and $Q_{\phi}$.

        \STATE \textcolor[RGB]{18,220,168}{\textbf{\# Start Policy Learning Thread:}}
        \STATE \textbf{Wait until} $\mathcal{R}$ contains at least 100 transitions.
        \FOR{each training step}
            \STATE Sample batch $b_{\text{demo}} \in \mathcal{D}_{E}$ and $b_{\text{online}} \in \mathcal{R}$.
            \STATE Update $\theta, \phi$ using $b_{\text{demo}} \cup b_{\text{online}}$.
        \ENDFOR

        \STATE \textcolor[RGB]{26,205,230}{\textbf{\# Start Interaction Thread:}}
        \FOR{each interaction step $t$}
            \STATE Compute deviation $\lambda_t \gets \text{FLOAT}(s_t, \mathcal{D}_{E})$.
            \IF{$\lambda_t \le \Lambda$} 
                \STATE Sample action $a_t \sim \pi_{\theta}(\cdot|s_t)$.
            \ELSE 
                \STATE \textcolor{gray}{\# Agent Intervention} \\
                \STATE $\mu \gets \mathcal{A}_{\text{vlm}}.\text{DecideMode}(s_t)$.
                
                \IF{$\mu$ is \textbf{Exploration Pruning}}
                    \STATE $flag \gets \mathcal{A}_{\text{vlm}}.\text{CheckMemory}(\mathcal{M})$.
                    \IF{$flag$ is True} 
                        \STATE \textcolor{gray}{\# Reuse Memory} \\ Retrieve $\mathcal{C}_{\text{box}} \sim \mathcal{M}$.
                    \ELSE
                        \STATE \textcolor{gray}{\# New Constraint} \\ $P_{\text{world}} \gets \mathcal{A}_{\text{vlm}}.\text{Perceive}(s_t)$.
                        \STATE $\mathcal{C}_{\text{box}} \gets \text{Geometry}(P_{\text{world}})$.
                        \STATE Update Memory $\mathcal{M} \gets \mathcal{M} \cup \mathcal{C}_{\text{box}}$.
                    \ENDIF
                    \STATE Sample $a_t \sim \pi_{\theta}(\cdot|s_t)$ s.t. $a_t \in \mathcal{C}_{\text{box}}$.
                
                \ELSIF{$\mu$ is \textbf{Action Guidance}}
                    \STATE $P_{\text{world}} \gets \mathcal{A}_{\text{vlm}}.\text{Perceive}(s_t)$.
                    \STATE $\mathbf{p}_{\text{way}} \gets \mathcal{A}_{\text{vlm}}.\text{GenWaypoint}(s_t,P_{\text{world}})$.
                    \STATE Compute $a_t$ derived from $\mathbf{p}_{\text{way}}$.
                \ENDIF
            \ENDIF
            \STATE Execute $a_t$, observe $r_t, s_{t+1}$.
            \STATE Store $(s_t, a_t, r_t, s_{t+1})$ in $\mathcal{R}$.
        \ENDFOR
    \end{algorithmic}
    \label{algo:apgs}
\end{algorithm}

\subsection{Training details}
\label{sec:train-details}

\noindent\textbf{Hardware and Perception.} We use a Franka Research 3 arm with a parallel gripper. The perception system includes wrist-mounted RealSense D405 cameras ($848 \times 480$) and side-mounted RealSense D435 or L515 cameras ($640 \times 480$). Specifically, USB Insertion uses two wrist cameras, Chinese Knot Hanging adds a side-mounted L515, and Towel Folding uses a side-mounted D435. For policy training, images are resized to $256 \times 256$, while the AGPS agent processes full-resolution inputs.

\noindent\textbf{Policy and Control.} Following HIL-SERL, we employ the DrQ algorithm. The policy state comprises visual observations and proprioceptive states, including end-effector poses, twists, forces, torques, and gripper status. The action space is a 6-DoF delta pose for the downstream impedance controller, extended to a 7-DoF vector with a binary gripper command for grasping tasks. Both data collection and control operate at 10 Hz.

\noindent\textbf{Implementation Details.} To minimize inference latency, we deploy the VLM agent (Qwen3-VL-235B-A22B-Instruct \cite{Qwen3-VL}) on an 8-GPU NVIDIA A800 server using the vLLM framework \cite{kwon2023efficient}. The reward classifier employs a ResNet-10 model pretrained on ImageNet, fine-tuned on 200 positive and 1000 negative samples. For baselines, ConRFT uses the pretrained Octo \cite{team2024octo} policy. In the AGPS FLOAT module, the expert dataset contains 20 trajectories for USB Insertion and Chinese Knot Hanging, and 10 for Towel Folding.

\subsection{VLM Prompt Templates}
\label{sec:prompt}
We present the core VLM templates for Action Guidance and Exploration Pruning below. Standard JSON formatting instructions and examples are omitted for brevity.

\begin{tcolorbox}[colback=white,colframe=black,title=\textbf{Action Guidance Template},breakable,skin=enhanced,fontupper=\small]
\textbf{Role:} Recovery and curriculum controller for a real-world RL expert controlling a Franka arm.\

\textbf{Objective:} Produce a safe primitive sequence to help policy recovery/learning.

\textbf{Inputs:}
\begin{itemize}
\item \{Task\_description\}, \{Keypoints\}.
\item One RGBD image with labeled keypoints.
\item \{Available\_tools\}, \{allowed\_primitive\}.
\end{itemize}
\textbf{Stage Decomposition Heuristics:}
\begin{itemize}
\item \textit{Hanging:} Lift (for clearance) $\rightarrow$ Translate (above hook target) $\rightarrow$ Descend (to place items).
\item \textit{Insertion:} Lift (if collision risk) $\rightarrow$ Descend/Advance (from plug tip to socket). Complete within 2 steps.
\item \textit{Deformable (Folding):} Move above source corner $\rightarrow$ Pre-grasp/Grasp $\rightarrow$ Lift (clearance) $\rightarrow$ Translate above target corner $\rightarrow$ Release.
\end{itemize}
\end{tcolorbox}

\vspace{-1.5pt}

\begin{tcolorbox}[
    colback=white, 
    colframe=black, 
    title=\textbf{Exploration Pruning Template}, 
    breakable, 
    skin=enhanced,
    fontupper=\small
]
\textbf{Role:} Precise visual grounding model and RL expert.\\
\textbf{Objective:} Construct ONE unified axis-aligned 3D exploration bounding box (AABB) in the robot BASE frame.

\textbf{Inputs:}
\begin{itemize}
    \item \texttt{\{task\_description\}}
    \item RGBD image with keypoints and base world frame coordinates.
    \item \texttt{\{keypoints\}} and \texttt{\{global\_xyz\_bounds\}}.
\end{itemize}

\textbf{Available Tools:}
\begin{itemize}
    \item \texttt{exploration\_bbox\_tool(}\allowbreak\texttt{mode="two\_keypoint\_union")}: \\
    Merges two local AABBs (source and target) into ONE unified outer envelope.
    \item \texttt{exploration\_bbox\_tool(}\allowbreak\texttt{mode="direct\_bbox3d")}: \\
    Builds one local AABB around a single keypoint.
\end{itemize}

\textbf{Task Instructions \& Constraints:}
\begin{itemize}
    \item \textit{Folding:} Uses \texttt{two\_keypoint\_union}. Output margins for both source and target. The final AABB is the tight outer envelope covering the necessary region.
    \item \textit{Insertion:} Uses \texttt{direct\_bbox3d} on the target. Output margins must be tight (typically 0.01-0.015m for x/y) while fully containing the port.
    \item \textit{Hard Constraint:} All margins must be non-negative magnitudes in meters.
\end{itemize}
\end{tcolorbox}


\section*{ACKNOWLEDGMENT}
We sincerely thank Yu Li and Yishuai Cai for invaluable support in setting up the hardware platform.

\bibliographystyle{IEEEtran}
\bibliography{references}

\end{document}